\documentclass[sigconf]{acmart}
\usepackage{amsmath}
\usepackage{subfigure}

\def\BibTeX{{\rm B\kern-.05em{\sc i\kern-.025em b}\kern-.08emT\kern-.1667em\lower.7ex\hbox{E}\kern-.125emX}}

\copyrightyear{2019}
\acmYear{2019}

\renewcommand\footnotetextcopyrightpermission[1]{}
\begin{document}

\title{Gravity-Inspired Graph Autoencoders for~Directed~Link~Prediction}

\author{Guillaume Salha}
\authornote{Contact author: research@deezer.com}
\affiliation{
  \institution{Deezer Research \& Development}
  \institution{LIX, \'{E}cole Polytechnique}
}

\author{Stratis Limnios}
\affiliation{
  \institution{LIX, \'{E}cole Polytechnique}
}

\author{Romain Hennequin}
\affiliation{
  \institution{Deezer Research \& Development}
}

\author{Viet Anh Tran}
\affiliation{
  \institution{Deezer Research \& Development}
}

\author{Michalis Vazirgiannis}
\affiliation{
  \institution{LIX, \'{E}cole Polytechnique \& AUEB}
}

\renewcommand{\shortauthors}{G. Salha et al.}

\begin{abstract}
Graph autoencoders (AE) and variational autoencoders (VAE) recently emerged as powerful node embedding methods. In particular, graph AE and VAE were successfully leveraged to tackle the challenging link prediction problem, aiming to figure out whether some pairs of nodes from a graph are connected by unobserved edges. However, these models focus on undirected graphs and therefore ignore the potential direction of the link, which is limiting for numerous real-life applications. In this paper, we extend the graph AE and VAE frameworks to address link prediction in directed graphs. We present a new gravity-inspired decoder that can effectively reconstruct directed graphs from node embedding representations. We empirically evaluate our method on three different directed link prediction tasks, for which standard graph AE and VAE perform poorly. We achieve competitive results on three real-world graphs, outperforming several popular baselines. 
\end{abstract}

\ccsdesc[500]{Information systems~Social networks}
\ccsdesc[500]{Mathematics of computing~Graph algorithms}
\ccsdesc[500]{Computing methodologies~Learning latent representations}

\keywords{Directed Graphs, Autoencoders, Variational Autoencoders, Graph Representation Learning, Node Embedding, Link Prediction}

\maketitle

\section{Introduction}

Graphs are useful data structures to represent relations among items efficiently. Due to the proliferation of graph data \cite{zhang2018network, wu2019comprehensive}, a large variety of specific problems initiated significant research efforts from the Machine Learning community, aiming to extract relevant information from such structures. This includes node clustering~\cite{malliaros2013clustering}, influence maximization \cite{kempe2003maximizing}, graph generation \cite{simonovsky2018graphvae}, and link prediction, on which we focus in this paper.

Link prediction consists in inferring the existence of new relations or still unobserved interactions (i.e., new \textit{edges} in the graph) between pairs of entities (\textit{nodes}) based on observable links and on their properties \cite{liben2007link, wang2015link}. This challenging task has been widely studied and successfully applied to several domains. In biological networks, link prediction models were leveraged to predict new interactions between proteins \cite{kovacs2019network}. It is also present in our daily lives, suggesting people we may know but are not connected to in our social networks \cite{liben2007link,wang2015link, haghani2017systemic}. Besides, link prediction is closely related to numerous recommendation tasks \cite{li2014recommendation, zhao2016genre, berg2018matrixcomp}.

Link prediction has been historically addressed through graph mining heuristics via the construction of similarity indices between nodes, capturing the likelihood of their connection in the graph. The Adamic-Adar and Katz indices \cite{liben2007link}, reflecting neighborhood structure and node proximity, are famous examples of such similarity indices. More recently,
along with the increasing efforts to extend Deep Learning methods to graph structures \cite{scarselli2009, bruna2014, wu2019comprehensive}, these approaches have been outperformed by others, adopting the node embedding paradigm \cite{tang2015line, grover2016node2vec, zhang2018network}. In a nutshell, their strategy is to train graph neural networks to represent nodes as vectors in a low-dimensional vector space, also known as an \textit{embedding} space. Ideally, nodes with structural proximity in the graph should be close to each other in such a space. Therefore, one could use proximity measures such as inner products between vector representations to predict new unobserved links in the underlying graph. In this direction, the graph extensions of autoencoders (AE) \cite{Rumelhart1986,baldi2012autoencoders} and variational autoencoders (VAE) \cite{kingma2013vae, Tschannen2018recentVAE} recently appeared as state-of-the-art approaches for link prediction in numerous experimental analyses \cite{wang2016structural, kipf2016-2, pan2018arga,  tran2018multitask, salha2019}.

However, these models focus on \textit{undirected graphs} and therefore ignore the potential direction of the link. As explained in Section 2, a graph autoencoder predicting that node $i$ is connected to node $j$ will also predict that node $j$ is connected to node $i$, with the same probability. This is limiting for numerous real-life applications, as \textit{directed graphs} are ubiquitous. For instance, web graphs are made up of directed hyperlinks. In social networks such as Twitter, opinion leaders are usually followed by many users, but only a few connections are reciprocal. Moreover, directed graphs are efficient abstractions in many domains where data are not explicitly structured as graphs. For instance, on music streaming services such as Deezer and Spotify, the page providing information about an artist will usually recommend the $k$ most similar artists. Artist similarities can be represented in a graph, in which nodes are artists connected to their $k$ most similar neighbors. Such a graph would be directed: while Bob Marley might be among the most similar artists of a new unknown reggae band, it is unlikely that this band would appear among Bob Marley's top similar artists on his page.

Directed link prediction has been tackled through the computation of asymmetric measures \cite{yu2014link, garcia2015phd, schall2015link} and, recently, a few attempts at capturing asymmetric proximity when creating node embeddings were proposed \cite{miller2009nonparametric, ou2016asymmetric, zhou2017scalable}. However, the question of how to effectively reconstruct directed graphs from vector space representations to perform directed link prediction remains widely open. In particular, it is unclear how to extend graph AE and graph VAE to directed graphs, and to which extent the promising performances of these models on undirected graphs could also be achieved on directed link prediction tasks. We propose to address these questions in this paper, making the following contributions:
\begin{itemize}
    \item We present a new model to effectively learn node embedding representations from directed graphs using the graph AE and VAE frameworks. We draw inspiration from Newton's theory of universal gravitation to introduce a new decoder, able to reconstruct asymmetric relations from vector space node embeddings.
    \item We empirically evaluate our approach on three different directed link prediction tasks, for which standard graph AE and VAE perform poorly. We achieve competitive results on three real-world datasets, outperforming popular baselines. To the best of our knowledge, these are the first graph AE/VAE experiments on directed graphs.
    \item We publicly release our code\footnote{\href{https://github.com/deezer/gravity_graph_autoencoders}{https://github.com/deezer/gravity\_graph\_autoencoders}} for these experiments, for reproducibility and easier future usage.
\end{itemize}

This paper is organized as follows. In Section 2, we recall key concepts related to graph AE and VAE and explain why these models are unsuitable for directed link prediction. In Section 3, we introduce our gravity-inspired method to reconstruct directed graphs using graph AE or VAE, and effectively perform directed link prediction. We present and discuss our experimental analysis in Section 4 and conclude in Section 5.

\section{Preliminaries}

In this section, we provide an overview of graph AE, graph VAE, and their main applications to link prediction. In the following, we consider a graph $\mathcal{G} = (\mathcal{V},\mathcal{E})$ without self-loops, with $|\mathcal{V}| = n$ nodes and $|\mathcal{E}| = m$ edges that can be directed. We denote by $A$ the adjacency matrix of $\mathcal{G}$, that is either binary or weighted. Moreover, nodes can possibly have feature vectors of size $f$, gathered in an $n \times f$ matrix $X$. For featureless graphs, $X$ is the $n \times n$ identity~matrix~$I$.

\subsection{Graph Autoencoders}

Graph autoencoders \cite{kipf2016-2, wang2016structural} are a family of unsupervised models extending autoencoders \cite{Rumelhart1986,baldi2012autoencoders} to graph structures. Their goal is to learn a node embedding space, i.e., low-dimensional ``latent'' vector representations of nodes. A graph AE involves two components:
\begin{itemize}
    \item Firstly, an \textit{encoder} model assigns a latent vector $z_i$ of size $d$, with $d \ll n$, to each node $i$ of the graph. The $n \times d$ matrix $Z$ of all latent vectors $z_i$ is usually the output of a Graph Neural Network (GNN) processing $A$ and $X$, i.e., we have: $Z = \text{GNN}(A,X)$.
    \item Then, a \textit{decoder} model aims to reconstruct the adjacency matrix $A$ from $Z$, using another GNN or a simpler alternative. For instance, in \cite{kipf2016-2} and in several extensions of their model \cite{pan2018arga, salha2019}, decoding is obtained through inner products between latent vectors, along with a sigmoid activation $\sigma(x) = 1/(1 + e^{-x})$ or, if $A$ is weighted, some more complex thresholding. In other words, the larger the inner product $z_i^T z_j$, the more likely node $i$ and $j$ are connected in the graph, according to the model. Denoting $\hat{A}$ the reconstruction of $A$ from the decoder, we have: $\hat{A} = \sigma(ZZ^T)$.
\end{itemize}
The intuition behind autoencoders is the following: if, starting from latent vectors, the decoder can reconstruct an adjacency matrix $\hat{A}$ that is \textit{close} to the original one, then these representations should preserve some important characteristics from the graph structure. Graph AE are trained by minimizing the reconstruction loss $\|A-\hat{A}\|_F$ where $\|\cdot\|_F$ is the Frobenius matrix norm \cite{wang2016structural}, or alternatively a weighted cross entropy loss \cite{kipf2016-2}, by gradient descent \cite{goodfellow2016deep}.

\subsection{Graph Convolutional Networks}

Throughout this paper, as Kipf and Welling~\cite{kipf2016-2} and most subsequent works \cite{pan2018arga,grover2018graphite, salha2019, do2019matrix}, we assume that the GNN encoder is a Graph Convolutional Network (GCN) \cite{kipf2016-1}. In a GCN with $L$ layers, with $L \geq 2$ and $Z = H^{(L)}$, we have:
$$\begin{cases} H^{(0)} = X ;\\
H^{(l)} = \text{ReLU}(\tilde{A} H^{(l-1)} W^{(l-1)}) \text{ for } l \in \{1,...L-1\} ;\\
H^{(L)} = \tilde{A} H^{(L-1)} W^{(L-1)}.
\end{cases} $$
In the above equation, $\tilde{A}$ denotes some normalized version of $A$. As undirected graphs were considered in previous models, a popular choice was the symmetric normalization $\tilde{A} = D^{-1/2}(A + I) D^{-1/2},$
where $D$ is the diagonal degree matrix of $A + I$. In essence, for each layer $l$, we average the feature vectors from $H^{(l-1)}$ of the neighbors of a given node, together with its own feature information (hence, the $I$) and with a ReLU activation: $\text{ReLU}(x) = \max(x,0)$. Weight matrices $W^{(l)}$ are trained using gradient descent techniques.

We rely on GCN encoders for three main reasons: 1) consistency with previous efforts on graph AE, 2) capitalization on the previous successes of GCN-based graph AE (see Section 2.4), and, last but not least, 3) computation efficiency. Indeed, the evaluation of each GCN layer has a linear complexity w.r.t. the number of edges $m$~\cite{kipf2016-1}. Speed-up strategies to improve the training of GCNs were also proposed \cite{chen2018fastgcn, wu2019simplifying}. Nonetheless, the method we present in this article is not limited to GCN. It would still be valid for alternative encoders, e.g., for more complex encoders, such as a ChebNet \cite{defferrard2016} that sometimes empirically outperform GCNs \cite{salha2019}.

\subsection{Variational Graph Autoencoders}

Kipf and Welling \cite{kipf2016-2} introduced variational graph autoencoders (denoted VGAEs in their work) as a graph extension of VAE \cite{kingma2013vae}. While sharing the name \textit{autoencoder}, graph VAE models are actually based on quite different mathematical foundations. Specifically, Kipf and Welling \cite{kipf2016-2} assume a probabilistic model on the graph that involves some latent variables $z_i$ of length $d \ll n$ for each node $i \in \mathcal{V}$. Such vectors are the node representations in a low-dimensional embedding space $\mathcal{Z}$. Denoting by $Z$ the $n\times d$ matrix of all latent vectors, the authors define the inference model as follows:
$$q(Z|A,X) = \prod_{i=1}^n q(z_i|A,X) \hspace{3pt} \text{where} \hspace{3pt} q(z_i|A,X) = \mathcal{N}(z_i|\mu_i, \text{diag}(\sigma_i^2)).$$
The latent vectors $z_i$ themselves are random samples drawn from the learned distribution, and this inference step is referred to as the \textit{encoding} part of the graph VAE. We learn parameters of Gaussian distributions using two GCNs. In other words, $\mu$, the matrix of mean vectors $\mu_i$, is defined as $\mu = \text{GCN}_{\mu}(A,X)$. Likewise, $\log \sigma = \text{GCN}_{\sigma}(A,X)$.

Then, a generative model attempts to reconstruct $A$ using, as for graph AE, inner products between latent variables:
$$p(A|Z) = \prod_{i=1}^n \prod_{j=1}^n p(A_{ij}|z_i, z_j) \hspace{3pt} \text{where} \hspace{3pt}p(A_{ij} = 1|z_i, z_j) = \sigma(z_i^Tz_j).$$
As before, $\sigma(\cdot)$ is the sigmoid activation function. This is the \textit{decoding} part of the model. One optimizes GCN weights by maximizing a tractable variational lower bound (ELBO) of the model's likelihood: 
$$\mathcal{L} = \mathbb{E}_{q(Z|A,X)} \Big[\log
p(A|Z)\Big] - \mathcal{D}_{KL}(q(Z|A,X)||p(Z)),$$
with a Gaussian prior $p(Z) = \prod_i p(z_i) = \prod_i \mathcal{N}(z_i|0,I)$, using gradient descent and leveraging the \textit{reparameterization trick} \cite{kingma2013vae}. $\mathcal{D}_{KL}(\cdot, \cdot)$ is the Kullback-Leibler divergence \cite{kullback1951information}.

\subsection{Graph AE and VAE for Undirected Link Prediction}

Over the last three years, graph AE/VAE and their extensions have been successfully leveraged to tackle several challenging tasks, such as node clustering \cite{wang2017mgae, pan2018arga, salha2019}, recommendation from bipartite graphs \cite{berg2018matrixcomp, do2019matrix}, and graph generation, notably biologically plausible molecule generation from VAE-based generative models \cite{molecule1,molecule2,molecule3,simonovsky2018graphvae, nevae2018}. We refer to the references above for a broader overview, and focus on link prediction tasks in the remainder of this section.

Link prediction has been the main evaluation task for graph AE and VAE in the work of Kipf and Welling \cite{kipf2016-2} and in numerous extensions \cite{tran2018multitask, pan2018arga, grover2018graphite, salha2019}. In a nutshell, authors evaluate the global ability of their models to predict whether some pairs of nodes from an undirected graph are connected by unobserved edges, using the latent space representations of the nodes. More formally, in such a setting, autoencoders are usually trained on an incomplete version of the graph where a proportion of the edges, say 10\%, were randomly removed. Then, a test set is created, gathering these missing edges and the same number of randomly picked pairs of unconnected nodes. Authors evaluate the model's ability to identify the true edges (i.e., $A_{ij} = 1$ in the \textit{complete} adjacency matrix) from the fake ones ($A_{ij} = 0$) using the decoding of the latent vectors  $\hat{A}_{i,j} = \sigma(z_i^T z_j)$. In other words, they predict that nodes are connected when $\hat{A}_{i,j}$ is larger than some threshold. This is a binary classification task, typically assessed using the \textit{Area Under the Receiver Operating Characteristic (ROC) Curve (AUC)} or the \textit{Average Precision (AP)} scores. For such tasks, graph AE and VAE have been empirically proven to be competitive and often outperforming w.r.t. several popular node embeddings baselines, notably Laplacian eigenmaps \cite{belkin2003laplacian} and word2vec-like models such as DeepWalk \cite{perozzi2014deepwalk}, LINE \cite{tang2015line}, and node2vec \cite{grover2016node2vec}.

We point out that most of these experiments focus on medium-size graphs with a few thousand nodes and edges. This is due to the limiting $O(dn^2)$ quadratic time complexity of the inner product decoder, which involves the multiplication of the dense matrices $Z$ and $Z^T$. Salha et al.~\cite{salha2019} recently overcame this scalability issue and introduced a general framework for more scalable graph AE and VAE, leveraging graph degeneracy concepts \cite{malliaros2019}. They confirmed the competitive performance of graph AE and VAE for large-scale link prediction based on experiments on undirected graphs with up to millions of nodes and edges.

\subsection{Why do these models fail to perform Directed Link Prediction?}

At this stage, we recall that all previously mentioned works assume, either explicitly or implicitly, that the input graph is \textit{undirected}. By design, graph AE and VAE are not suitable for directed graphs, since they ignore directions when reconstructing the adjacency matrix from the embedding. Indeed, due to the symmetry of the inner product decoder, we have:
$$\hat{A}_{ij} = \sigma(z_i^Tz_j) = \sigma(z_j^Tz_i) = \hat{A}_{ji}.$$
In other words, if we predict the existence of an edge $(i,j)$ from node $i$ to node $j$, then we also necessarily predict the existence of the reverse edge $(j,i)$, with the same probability. Consequently, as shown in Section 4, standard graph AE and VAE significantly underperform on link prediction tasks in directed graphs, where relations are not always reciprocal.

Replacing inner product decoders by an $L_p$ distance in the embedding (e.g., the Euclidean distance, if $p=2$) or by existing more refined decoders \cite{grover2018graphite} would lead to the same conclusion, since they are also symmetric. Recently, Zhang et al.~\cite{zhang2019d} proposed D-VAE, a variational autoencoder for small Directed Acyclic Graphs (DAG) such as neural network architectures or Bayesian networks, focusing on neural architecture search and structure learning. However, the question of how to extend graph AE and VAE to general directed graphs, such as citation networks or web hyperlink networks where directed link prediction is challenging, remains open.

\subsection{On the Source/Target Vectors Paradigm}
To conclude these preliminaries, we highlight that, out of the graph AE/VAE frameworks, a few recent node embedding methods proposed to tackle directed link prediction by actually learning \textit{two latent vectors} for each node. More precisely: 
\begin{itemize}
    \item HOPE, short for \textit{High-Order Proximity preserved Embedding} \cite{ou2016asymmetric}, aims to preserve high-order node proximities and capture asymmetric transitivity. Nodes are represented by \textit{two} vectors: \textit{source} vectors $z^{(s)}_i$, stacked up in an $n \times d$ matrix $Z^{(s)}$, and \textit{target} vectors $z^{(t)}_i$, gathered in another $n \times d$ matrix $Z^{(t)}$. For a given $n \times n$ similarity matrix $S$, authors learn these vectors by approximately minimizing $ \|S - Z^{(s)} Z^{(t) T} \|_F$ using a generalized SVD. For directed graphs, an usual choice for $S$ is the Katz matrix $S^{\text{Katz}} = \sum_{i=1}^{\infty} \beta^ i A^i$, with $S^{\text{Katz}} = (I - \beta A)^{-1} \beta A$ if the parameter $\beta > 0$ is smaller than the spectral radius of $A$ \cite{katz1953new}. It computes the number of paths from a node to another one, these paths being exponentially weighted according to their length. For link prediction, one can assess the likelihood of a link from node $i$ to node $j$ using the asymmetric reconstruction $\hat{A}_{ij} = \sigma(z^{(s)T}_i z^{(t)}_j)$.
    \item APP \cite{zhou2017scalable} is a scalable \textit{Asymmetric Proximity Preserving} node embedding method, that preserves the Rooted PageRank score \cite{page1999pagerank} for any node pair. APP leverages random walk with restart strategies to learn, like HOPE, a source vector and a target vector for each node. As before, it predicts that node $i$ is connected to node $j$ from the inner product of source vector $i$ and target vector $j$, with a sigmoid activation.
\end{itemize}

One can derive a straightforward extension of this \textit{source/target vectors} paradigm for graph AE and VAE. Indeed, considering GCN encoders returning $d$-dimensional latent vectors $z_i$, with $d$ being even, we can assume that the $d/2$ first dimensions (resp. the $d/2$ last dimensions) of $z_i$ actually correspond to the source (resp. target) vector of node $i$, i.e., $z^{(s)}_i = z_{i [1:\frac{d}{2}]}$ and $z^{(t)}_i = z_{i [(\frac{d}{2}+1):d]}$. Then, we can replace the symmetric decoder $\hat{A}_{ij} = \hat{A}_{ji} = \sigma(z^T_i z_j)$ by $\hat{A}_{ij} = \sigma(z^{(s)T}_i z^{(t)}_j)$ and $\hat{A}_{ji} = \sigma(z^{(s)T}_j z^{(t)}_i)$,
both in the AE and VAE frameworks, to reconstruct directed links from the encoded representations. We refer to this method as \textit{source/target graph AE (or VAE)} in this paper.

However, in the following, we adopt a different approach, and we propose to come back to the original idea consisting in learning a \textit{single} node embedding space, and therefore represent each node via a single latent vector. Such an approach has a stronger interpretability power. As we later show in the experimental part of this paper, it also outperforms source/target graph AE and VAE on directed link prediction tasks.

\section{A Gravity-inspired model for Directed Graph AE and VAE}

This section introduces a new model to learn node embedding representations from directed graphs using graph AE and VAE, and to address directed link prediction problems. The main challenge is the following: how to effectively reconstruct asymmetric relations from representations that are (unique) latent vectors in a node embedding space where inner products and distances~are~symmetric?

To overcome this challenge, we resort to classical mechanics and especially Newton's theory of universal gravitation. We propose an analogy between latent node representations in an embedding space and celestial objects in space. Specifically, although the Earth-Moon distance is symmetric, the \textit{acceleration} of the Moon towards the Earth due to gravity is larger than the acceleration of the Earth towards the Moon. As explained below, this is due to the fact that the Earth is more massive than the Moon. In this Section 3, we transpose these notions of mass and acceleration to node embeddings to build up our asymmetric gravity-inspired decoder.

\subsection{Newton's Theory of Universal Gravitation}

According to Newton's theory of universal gravitation \cite{newton1687philosophiae}, each particle in the universe attracts the other particles through a force called \textit{gravity}. This force is proportional to the product of the masses of the particles, and inversely proportional to the squared distance between their centers. More formally, let us denote by $m_1$ and $m_2$ the positive masses of two objects $1$ and $2$ and by $r$ the distance between their centers. Then, the gravitational force $F$ attracting the two objects is:
$$F = \frac{Gm_1m_2}{r^2},$$
where $G$ is the gravitational constant \cite{cavendish1798xxi}. Then, using Newton's second law of motion \cite{newton1687philosophiae}, we derive $a_{1 \rightarrow 2}$, the acceleration of object $1$ towards object $2$ due to gravity:
$$a_{1 \rightarrow 2} =  \frac{F}{m_1} = \frac{Gm_2}{r^2}.$$
Likewise, the acceleration $a_{2 \rightarrow 1}$ of $2$ towards $1$ due to gravity is:
$$a_{2 \rightarrow 1} = \frac{F}{m_2} = \frac{Gm_1}{r^2}.$$
We note that $a_{1 \rightarrow 2} \neq a_{2 \rightarrow 1}$ when $m_1 \neq m_2$. More precisely, we have $a_{1 \rightarrow 2} > a_{2 \rightarrow 1}$ when $m_2 > m_1$ and conversely, i.e., the acceleration of the less massive object towards the more massive object due to gravity is higher.

Despite being superseded in modern physics by Einstein's theory of general relativity \cite{einstein1915erklarung}, describing gravity not as a force but as a consequence of spacetime curvature, Newton's law of universal gravitation is still used in many applications, as the theory provides precise approximations of the effect of gravity when gravitational fields are not extreme. In the remainder of this paper, we draw inspiration from this theory, notably from the formulation of acceleration, to build our proposed autoencoder models. We highlight that Newtonian gravity concepts were already successfully leveraged in \cite{gravity1} for graph visualization, and in \cite{gravity2} where the force formula has been applied through graph mining measures to construct symmetric similarity scores between nodes.

\subsection{From Physics to Node Representations} We come back to our initial analogy between celestial objects in space and node embedding representations. In this Section 3.2, we assume that, in addition to a latent vector $z_i$ of dimension $d \ll n$, we have at our disposal a model that is also able to learn a new \textit{mass parameter} $m_i \in \mathbb{R}^+$ for each node $i \in \mathcal{V}$ of a directed graph. Such a parameter would capture the propensity of $i$ to attract other nodes from its neighborhood in this graph, i.e., to make them point towards $i$ through a directed edge. We could apply Newton's equations in the embedding space resulting from such an augmented model. Specifically, we could use the \textit{acceleration} $a_{i \rightarrow j} = \frac{G m_j}{r^2}$ of a node $i$ towards a node $j$ due to gravity in the embedding as an indicator of the likelihood that $i$ is connected to $j$ in the directed graph, with $r^2 = \|z_i - z_j\|_2^2$. In a nutshell:
\begin{itemize}
    \item The numerator would capture that some nodes are more influential than others. For instance, in a scientific publications citation network, groundbreaking articles are more influential and should be more cited. Here, the bigger $m_j$ the more likely $i$ is connected to $j$ via the $(i,j)$ directed edge.
    \item The denominator would highlight that nodes with structural proximity in the graph, typically with a common neighborhood, are more likely to be connected, provided that the model effectively manages to embed these nodes \textit{close} to each other in the embedding space. For instance, in a scientific publications citation network, the article $i$ will more likely cite the article $j$ if it comes from a similar field~of~study.
\end{itemize}

However, instead of directly dealing with $a_{i \rightarrow j}$, we choose to use $\log a_{i \rightarrow j}$ in the remainder of this paper. This logarithmic transform has two advantages. Firstly, its concavity limits the potentially large values resulting from the acceleration towards very central nodes. Also, $\log a_{i \rightarrow j}$ can be negative, which is more convenient to reconstruct an unweighted edge (i.e., in the adjacency matrix $A$ we have $A_{ij} = 1$ or $0$) using a sigmoid activation function, as follows:
\begin{align*}
\hat{A}_{ij} &= \sigma(\log a_{i \rightarrow j}) \\
&= \sigma(\underbrace{\log G m_j}_{\tilde{m}_j} - \log \|z_i - z_j\|_2^2)
\end{align*}

\subsection{Gravity-Inspired Directed Graph AE}

For pedagogical purposes, we assumed in Section 3.2 that we had a model able to learn mass parameters $m_i$ for all $i \in \mathcal{V}$. We now explain how to derive such parameters using the graph autoencoder framework.

\subsubsection{Encoder}

For the encoder part of the model, we leverage a Graph Convolutional Network processing $A$ and, potentially, a node features matrix $X$. Such a GCN assigns a vector of size $(d+1)$ to each node of the graph, instead of $d$ as in standard graph autoencoders. The first $d$ dimensions correspond to the latent vector representation of the node, i.e., $z_i$, where $d \ll n$ is the dimension of the latent vectors in the node embedding. The last value of the output vector is the model's estimate of $\tilde{m}_i = \log G m_i$. To sum up, we have:
$$(Z,\tilde{M}) = GCN(A,X),$$
where $Z$ is the $n \times d$ matrix of all latent vectors $z_i$, $\tilde{M}$ is the $n$-dimensional vector of all values of $\tilde{m}_i$, and $(Z,\tilde{M})$ is the $n \times (d+1)$ matrix row-concatenating $Z$ and $\tilde{M}$. We note that learning $\tilde{m}_i$ is equivalent to learning $m_i$, but is also more convenient since we get rid of the gravitational constant $G$ and of the logarithm.

In this GCN encoder, as we process directed graphs, we replace the usual symmetric normalization of $A$, i.e., $D^{-1/2}(A + I) D^{-1/2}$, by the out-degree normalization $D_{\text{out}}^{-1} (A + I)$. Here, $D_{\text{out}}$ denotes the diagonal out-degree matrix of $A + I$, i.e., the element $(i,i)$ of $D_{\text{out}}$ corresponds to the number of edges (potentially weighted) going out of the node $i$, plus one. Therefore, at each layer of the GCN, the feature vector of a node is the average of feature vectors from the previous layer of the neighbors to which it points, together with its own feature vector and combined with a ReLU activation.

\subsubsection{Decoder} We leverage the previously defined logarithmic version of the acceleration, together with a sigmoid activation, to reconstruct the adjacency matrix $A$ from $Z$ and $\tilde{M}$. Denoting $\hat{A}$ the reconstruction of $A$, we have:
$$\hat{A}_{ij} = \sigma(\tilde{m}_j - \log \|z_i - z_j\|^2_2).$$
Contrary to the inner product decoder, we usually have $\hat{A}_{ij} \neq \hat{A}_{ji}$. This approach is, therefore, relevant for directed graph reconstruction. During training, we aim to minimize the reconstruction loss from matrix $A$, formulated as a weighted cross entropy loss as in \cite{kipf2016-2}, by gradient descent.

\subsection{Gravity-Inspired Directed Graph VAE}

We also propose to extend our gravity-inspired method to the graph variational autoencoder framework.

\subsubsection{Encoder} 

We extend \cite{kipf2016-2} to build up an inference model for $(Z,\tilde{M})$. In other words, the $(d+1)$-dimensional latent vector associated with each node $i$ is $(z_i,\tilde{m}_i)$, concatenating the $d$-dimensional vector $z_i$ and the scalar $\tilde{m}_i$. We have:
$$q((Z,\tilde{M})|A,X) = \prod_{i=1}^n q((z_i,\tilde{m}_i)|A,X),$$
with Gaussian hypotheses on the distributions, as in \cite{kipf2016-2}: 
$$q((z_i,\tilde{m}_i)|A,X) = \mathcal{N}((z_i,\tilde{m}_i)|\mu_i, \text{diag}(\sigma_i^2)).$$
Parameters of Gaussian distributions are learned using two GCNs, with a similar out-degree normalization w.r.t. Section 3.3:
$$\mu = \text{GCN}_{\mu}(A,X) \hspace{5pt} \text{and} \hspace{5pt} \log \sigma = \text{GCN}_{\sigma}(A,X).$$

\subsubsection{Decoder}

Using vectors $(z_i,\tilde{m}_i)$ sampled from these distributions, we incorporate our gravity-inspired decoding scheme into the graph VAE generative model, attempting to reconstruct $A$:
$$p(A|Z,\tilde{M}) = \prod_{i=1}^n \prod_{j=1}^n p(A_{ij}|z_i, z_j, \tilde{m}_j),$$
where:
$$p(A_{ij} = 1|z_i, z_j, \tilde{m}_j) = \sigma(\tilde{m}_j - \log \|z_i - z_j\|^2_2).$$
As \cite{kipf2016-2}, we train the model by maximizing the ELBO of the model's likelihood using gradient descent techniques and with a Gaussian prior $p((Z,\tilde{M})) = \prod_i p(z_i,m_i) = \prod_i \mathcal{N}((z_i,m_i)|0,I)$. We discuss these Gaussian assumptions in the experimental part of this paper.

\subsection{Generalization of the Decoder}

We point out that one can improve the flexibility of our decoder, both in the AE and VAE settings, by introducing an additional parameter $\lambda \in \mathbb{R}^+$ and reconstruct $\hat{A}_{ij}$ as follows:
$$\hat{A}_{ij} = \sigma(\tilde{m}_j - \lambda \log \|z_i - z_j\|^2_2).$$
Decoders from Sections 3.3 and 3.4 are special cases where $\lambda = 1$. This parameter can be tuned by cross-validation on link prediction tasks (see Section 4). The interpretation of such a parameter is twofold. Firstly, it constitutes a simple tool to balance the relative importance of the node distance in the embedding for reconstruction w.r.t. the mass attraction parameter. Then, from a physical point of view, it is equivalent to replacing the squared distance in Newton's formula with a distance to the power of $2\lambda$. Our experimental analysis on link prediction provides insights on when and why deviating from Newton's theory (i.e., $\lambda = 1$) is relevant.

\subsection{On Complexity and Scalability}

Assuming featureless nodes, a sparse representation of $A$ with $m$ non-zero entries, and considering that our models return an $n \times (d+1)$ dense matrix, then the space complexity of our approach is $O(m + n(d+1))$, both in the AE and VAE frameworks. If nodes also have features (in the $n \times f$ matrix $X$), then the space complexity becomes $O(m + n(f+d+1))$, with $d \ll n$ and $f \ll n$ in practice. Therefore, as standard graph AE and VAE models \cite{kipf2016-2}, space complexity increases linearly w.r.t. the size of the graph.

Moreover, due to the pairwise computations of $L_2$ distances between all  $d$-dimensional vectors $z_i$ and $z_j$ involved in our gravity-inspired decoder, our models have a quadratic time complexity $O(dn^2)$ w.r.t. the number of nodes in the graph, as standard graph AE and VAE models. Consequently, in our experiments we focus on medium-size datasets, i.e., graphs with thousands of nodes and edges. We nevertheless point out that extending our model to large graphs (with millions of nodes and edges) could be achieved by applying the degeneracy framework proposed in \cite{salha2019} to scale graph autoencoders, or a variant of their approach involving directed graph degeneracy concepts \cite{giatsidis2013d}. We will provide a deeper investigation of these scalability concerns in future work.

\section{Experimental Analysis}

In this section, we empirically evaluate and discuss the performance of our models, on three real-world datasets and three variants of the directed link prediction problem.

\subsection{Three Directed Link Prediction Tasks}

We consider the following three learning tasks for our experiments.

\subsubsection{Task 1: General Directed Link Prediction}

The first task is referred to as \textit{general directed link prediction}. As in previous works \cite{kipf2016-1, grover2018graphite, pan2018arga, salha2019}, we train models on incomplete versions of graphs where $15\%$ of edges were randomly removed. We take directionality into account in the masking process. In other words, if a link between node $i$ and $j$ is reciprocal, we can possibly remove the $(i,j)$ edge but still observe the reverse $(j,i)$ edge in the incomplete training graph. Then, we create some validation and test sets from the removed edges and from the same number of randomly sampled pairs of unconnected nodes. We evaluate the performance of our models on a binary classification task consisting in identifying the actually removed edges from the fake ones, and evaluating results using the AUC and AP scores. In the following, the validation set contains $5\%$
of edges, and the test set includes $10\%$ of edges. The validation set is only used for hyperparameter tuning.

This setting corresponds to the most general formulation of link prediction. However, due to the large number of unconnected pairs of nodes in numerous real-world graphs, we expect the impact of directionality on performances to be limited. Indeed, for each unidirectional edge $(i,j)$ from the graph, it is unlikely to retrieve the reverse (unconnected) pair $(j,i)$ among negative samples in the test set. Consequently, models focusing on graph proximity and ignoring the direction of the link, such as standard graph AE and VAE, might still perform reasonably well on such a task.

For this reason, in the remainder of this Section 4.1, we also propose two additional learning tasks designed to reinforce the importance of directionality learning.

\subsubsection{Task 2: Biased Negative Samples (BNS.) Link Prediction}

For the second task, we also train models on incomplete versions of graphs where $15\%$ of edges were removed: $5\%$ for the validation set and $10\%$ for the test set. However, removed edges are all \textit{unidirectional}, i.e., $(i,j)$ exists but not $(j,i)$. The reverse node pairs are included in the validation and test sets in this setting. They constitute negative samples. In other words, all node pairs from validation and test sets are included in \textit{both} directions. As for \textit{general directed link prediction} task, we evaluate the performance of our models on a binary classification task consisting in identifying actual edges from fake ones, and therefore evaluate the ability of our models to correctly reconstruct $A_{ij} = 1$ and $A_{ji} = 0$ \textit{simultaneously}.

This task has been presented in \cite{zhou2017scalable} under the name \textit{biased negative samples link prediction}. It is more challenging than general link direction, as the ability to reconstruct asymmetric relations is now crucial. Models ignoring directionality and only learning the symmetric graph proximity, such as standard graph AE and VAE, will fail in such a setting (they would always return $\hat{A}_{ij} = \hat{A}_{ji}$).

\subsubsection{Task 3: Bidirectionality Prediction}

As a third task, we evaluate the ability of our models to identify \textit{bidirectional} edges, i.e., reciprocal connections, from \textit{unidirectional} edges. We create an incomplete training graph by randomly removing one of the two directions of all bidirectional edges. Therefore, the training graph only includes unidirectional connections. Then, we again consider a binary classification problem. We aim to retrieve bidirectional edges in a test set composed of their removed direction and of the same number of reverse directions from true unidirectional edges (that are, therefore, fake edges). In other words, for each pair of nodes $i,j$ from the test set, we observe a connection from $j$ to $i$ in the incomplete training graph, but only half of them are reciprocal. This third evaluation task, referred to as \textit{bidirectionality prediction} in this paper, also strongly relies on directionality learning. Consequently, as for task 2, standard graph AE and VAE are expected to~perform~poorly.

\subsection{Experimental Setting}

\begin{table}[t]
\begin{center}
\begin{small}
\centering
\caption{Directed graphs used in our experiments.\label{tab:Link_prediction datasets}}
\begin{tabular}{c|c|c|c}
\toprule
\textbf{Dataset} & \textbf{Number of} & \textbf{Number of} & \textbf{Percentage of} \\
& \textbf{nodes} & \textbf{edges} & \textbf{reciprocity} \\
\midrule
\midrule
\textbf{Cora} & $2\, 708$ & $5\, 429$ & $2.86 \%$ \\
\textbf{Citeseer} & $3\, 327$ & $4\, 732$ & $1.20\%$ \\
\textbf{Google} & $15\, 763$ & $171\, 206$ & $14.55\%$ \\
\bottomrule
\end{tabular}
\end{small}
\end{center}
\vspace{-0.5cm}
\end{table}

\subsubsection{Datasets}

\begin{table*}[ht]
  \centering
\begin{footnotesize}
  \caption{Directed link prediction on the Cora, Citeseer and Google graphs.}
  \begin{tabular}{c|c|cc|cc|cc}
    \toprule
    \textbf{Dataset}  & \textbf{Model} & \multicolumn{2}{c}{\textbf{Task 1: General Link Prediction}} & \multicolumn{2}{c}{\textbf{Task 2: B.N.S. Link Prediction}} & \multicolumn{2}{c}{\textbf{Task 3: Bidirectionality Prediction}} \\
    &  & \tiny \textbf{AUC (in \%)} & \tiny \textbf{AP (in \%)} & \tiny \textbf{AUC (in \%)} & \tiny \textbf{AP (in \%)} & \tiny \textbf{AUC (in \%)} & \tiny \textbf{AP (in \%)}\\
    \midrule
    \midrule
    \textbf{Cora} & \textit{Gravity Graph VAE (ours)} & $91.92 \pm 0.75$ & $\textbf{92.46} \pm \textbf{0.64}$ & $\textbf{83.33} \pm \textbf{1.11}$ & $\textbf{84.50} \pm \textbf{1.24}$ & $\textbf{75.00} \pm \textbf{2.10}$ & $\textbf{73.87} \pm \textbf{2.82}$ \\
     & \textit{Gravity Graph AE (ours)} & $87.79 \pm 1.07$ & $90.78 \pm 0.82$ & $\textbf{83.18} \pm \textbf{1.12}$ & $\textbf{84.09} \pm \textbf{1.16}$ & $\textbf{75.57} \pm \textbf{1.90}$ & $\textbf{73.40} \pm \textbf{2.53}$ \\
     \cmidrule{2-8}
     & Standard Graph VAE & $82.79 \pm 1.20$ & $86.69 \pm 1.08$ & $50.00 \pm 0.00$ & $50.00 \pm 0.00$ & $58.12 \pm 2.62$ & $59.70 \pm 2.08$ \\
     & Standard Graph AE & $81.34 \pm 1.47$ & $82.10 \pm 1.46$ & $50.00 \pm 0.00$ & $50.00 \pm 0.00$ & $53.07 \pm 3.09$ & $54.60 \pm 3.13$ \\
     & Source/Target Graph VAE & $85.34 \pm 1.29$ & $88.35 \pm 0.99$ & $63.00 \pm 1.05$ & $64.62 \pm 1.37$ & $\textbf{75.20} \pm \textbf{2.62}$ & $\textbf{73.86} \pm \textbf{3.04}$ \\
     & Source/Target Graph AE & $82.67 \pm 1.42$ & $83.25 \pm 1.51$ & $57.81 \pm 2.64$ & $57.66 \pm 3.35$ & $65.83 \pm 3.87$ & $63.15 \pm 4.58$ \\
     & APP & $\textbf{93.92} \pm \textbf{1.01}$ & $\textbf{93.26} \pm \textbf{0.60}$ & $69.20 \pm 0.65$ & $67.93 \pm 1.09$ & $72.85 \pm 1.91$ & $70.97 \pm 2.60$ \\
     & HOPE & $80.82 \pm 1.63$ & $81.61 \pm 1.08$  & $61.84 \pm 1.84$ & $63.73 \pm 1.12$ & $65.11 \pm 1.40$ & $64.24 \pm 1.18$ \\
     & node2vec & $79.01 \pm 2.00$ & $84.20 \pm 1.62$ & $50.00 \pm 0.00$ & $50.00 \pm 0.00$ & $66.97 \pm 1.41$ & $67.61 \pm 1.80$ \\
    \midrule
    \midrule
    \textbf{Citeseer} & \textit{Gravity Graph VAE (ours)} & $\textbf{87.67} \pm \textbf{1.07}$ & $\textbf{89.79} \pm \textbf{1.01}$ & $\textbf{76.19} \pm \textbf{1.35}$ & $\textbf{79.27} \pm \textbf{1.24}$ & $\textbf{71.61} \pm \textbf{3.20}$ & $\textbf{71.87} \pm \textbf{3.87}$ \\
     & \textit{Gravity Graph AE  (ours)} & $78.36 \pm 1.55$ & $84.75 \pm 1.10$ & $\textbf{75.32} \pm \textbf{1.53}$ & $\textbf{78.47} \pm \textbf{1.27}$ & $\textbf{71.48} \pm \textbf{3.64}$ & $\textbf{71.50} \pm \textbf{3.62}$ \\
     \cmidrule{2-8}
     & Standard Graph VAE & $78.56 \pm 1.43$ & $83.66 \pm 1.09$ & $50.00 \pm 0.00$ & $50.00 \pm 0.00$ & $47.66 \pm 3.73$ & $50.31 \pm 3.27$ \\
     & Standard Graph AE & $75.23 \pm 2.13$ & $75.16 \pm 2.04$ & $50.00 \pm 0.00$ & $50.00 \pm 0.00$ & $45.01 \pm 3.75$ & $49.79 \pm 3.71$ \\
     & Source/Target Graph VAE & $79.45 \pm 1.75$ & $83.66 \pm 1.32$ & $57.32 \pm 0.92$ & $61.02 \pm 1.37$ & $69.67 \pm 3.12$ & $67.05 \pm 4.10$ \\
     & Source/Target Graph AE & $73.97 \pm 3.11$ & $75.03 \pm 3.37$ & $56.97 \pm 1.33$ & $57.62 \pm 2.62$ & $54.88 \pm 6.02$ & $55.81 \pm 4.93$ \\
     & APP & $\textbf{88.70} \pm \textbf{0.92}$ & $\textbf{90.29} \pm \textbf{0.71}$ & $64.35 \pm 0.45$ & $63.70 \pm 0.51$ & $64.16 \pm 1.90$ & $63.77 \pm 3.28$ \\
     & HOPE & $72.91 \pm 0.59$ & $71.29 \pm 0.52$ & $60.24 \pm 0.51$ & $61.28 \pm 0.57$ & $52.65 \pm 3.05$ & $54.87 \pm 1.67$ \\
     & node2vec & $71.02 \pm 1.78$  & $77.70 \pm 1.22$ & $50.00 \pm 0.00$ & $50.00 \pm 0.00$ & $61.08 \pm 1.88$ & $63.63 \pm 2.77$ \\
    \midrule
    \midrule
    \textbf{Google} & \textit{Gravity Graph VAE (ours)} & $\textbf{97.84} \pm \textbf{0.25}$ & $\textbf{98.18} \pm \textbf{0.14}$ & $\textbf{88.03} \pm \textbf{0.25}$ & $\textbf{91.04} \pm \textbf{0.14}$ & $84.69 \pm 0.31$ & $84.89 \pm 0.30$ \\
     & \textit{Gravity Graph AE (ours)} & $\textbf{97.77} \pm \textbf{0.10}$ & $\textbf{98.43} \pm \textbf{0.10}$ & $\textbf{87.71} \pm \textbf{0.29}$ & $\textbf{90.84} \pm \textbf{0.16}$ & $\textbf{85.82} \pm \textbf{0.63}$ & $\textbf{85.91} \pm \textbf{0.50}$ \\
     \cmidrule{2-8}
     & Standard Graph VAE & $87.14 \pm 1.20$ & $88.14 \pm 0.98$ & $50.00 \pm 0.00$ & $50.00 \pm 0.00$ & $40.03 \pm 4.98$ & $44.69 \pm 3.52$ \\
     & Standard Graph AE & $91.34 \pm 1.13$ & $92.61 \pm 1.14$ & $50.00 \pm 0.00$ & $50.00 \pm 0.00$ & $41.35 \pm 1.92$ & $41.92 \pm 0.81$ \\
     & Source/Target Graph VAE & $96.33 \pm 1.04$ & $96.24 \pm 1.06$ & $85.30 \pm 3.18$ & $84.69 \pm 4.42$ & $75.11 \pm 2.07$ & $73.63 \pm 2.06$ \\
     & Source/Target Graph AE & $\textbf{97.76} \pm \textbf{0.41}$ & $\textbf{97.74} \pm \textbf{0.40}$ & $86.16 \pm 2.95$ & $86.26 \pm 3.33$ & $82.27 \pm 1.29$ & $80.10 \pm 1.80$ \\
     & APP & $97.04 \pm 0.10$ & $96.97 \pm 0.11$ & $83.06 \pm 0.46$ & $85.15 \pm 0.42$ & $73.43 \pm 0.16$ & $68.74 \pm 0.19$ \\
     & HOPE & $81.16 \pm 0.67$ & $83.02 \pm 0.35$ & $74.23 \pm 0.80$ & $72.70 \pm 0.79$ & $70.45 \pm 0.18$ & $70.84 \pm 0.22$ \\
     & node2vec & $83.11 \pm 0.27$ & $85.79 \pm 0.30$ & $50.00 \pm 0.00$ & $50.00 \pm 0.00$ & $78.99 \pm 0.35$ & $76.72 \pm 0.53$ \\
    \bottomrule
  \end{tabular}
  \end{footnotesize}
\end{table*} 
We provide experiments on three publicly available real-world directed graphs, whose statistics are presented in Table~1. The \textit{Cora}\footnote{\href{https://linqs.soe.ucsc.edu/data}{https://linqs.soe.ucsc.edu/data}\label{linqs}} and \textit{Citeseer}\textsuperscript{\ref{linqs}} datasets are \textit{citation graphs} consisting of scientific publications citing one another. The \textit{Google}\footnote{\href{http://konect.uni-koblenz.de/networks/cfinder-google}{http://konect.uni-koblenz.de/networks/cfinder-google}} dataset is a \textit{web graph}, whose nodes are web pages and directed edges represent hyperlinks between them. The \textit{Google} graph is denser than \textit{Cora} and \textit{Citeseer} and has a higher proportion of bidirectional edges. Graphs are unweighted and featureless.

\subsubsection{Standard and Gravity-Inspired Autoencoders}

We train gravity-inspired AE and VAE models for each graph. For comparison purposes, we also train standard graph AE and VAE from \cite{kipf2016-2}.
Each of these four models includes a two-layer GCN encoder with a 64-dimensional hidden layer and with an out-degree left normalization of $A$ as defined in Section 3.3.1. All models are trained for 200 epochs and return 32-dimensional latent vectors. We use the Adam optimizer \cite{kingma2014adam}, apply a learning rate of 0.1 for \textit{Cora} and \textit{Citeseer} and 0.2 for \textit{Google}, train models without dropout, performing full-batch gradient descent and using the reparameterization trick \cite{kingma2013vae} for variational autoencoders. Also, for \textit{tasks 1 and 3}, we picked $\lambda = 1$ (respectively $\lambda = 10$)  for \textit{Cora} and \textit{Citeseer} (resp. for \textit{Google}) ; for \textit{task 2} we picked $\lambda = 0.05$ for all three graphs, which we interpret in the next sections. All hyperparameters were tuned from the AUC score on \textit{task 1}, i.e., on general directed link prediction task.

\subsubsection{Baselines}

Besides comparing to standard graph AE and VAE models, we also compare the performance of our methods to the alternative graph embedding methods mentioned in Section 2.6:
\begin{itemize}
    \item Our own \textit{source/target graph AE and VAE}, extending the source/target vectors paradigm to graph AE/VAE, and trained with similar settings w.r.t. standard and gravity models.
    \item HOPE \cite{ou2016asymmetric}, with $\beta = 0.01$ and source and target vectors of dimension 16, to learn 32-dimensional node representations.
    \item APP \cite{zhou2017scalable}, training models over 100 iterations to learn 16-dimensional source and target vectors, i.e., 32-dimensional node representations, using the implementation from \cite{zhou2017scalable}.
    \item For comparison purposes, in our experiments, we also consider \textit{node2vec} models \cite{grover2016node2vec} that, while dealing with the directionality in random walks, only return a 32-dimensional embedding vector per node. We rely on \textit{symmetric} inner products with sigmoid activation for link prediction, and we, therefore, expect node2vec to underperform on tasks 2 and 3. We trained models from 10 random walks of length 80 per node, with $p = q =1$ and a window size of 5.
\end{itemize}
We used Python and especially the Tensorflow library, except for APP, where we used the authors' Java implementation \cite{zhou2017scalable}. We trained models on an NVIDIA GTX 1080 GPU and ran other operations on a double Intel Xeon Gold 6134 CPU.

\subsection{Results for Directed Link Prediction}

 \begin{figure*}[t]
 	\centering
 \includegraphics[width=.76\textwidth]{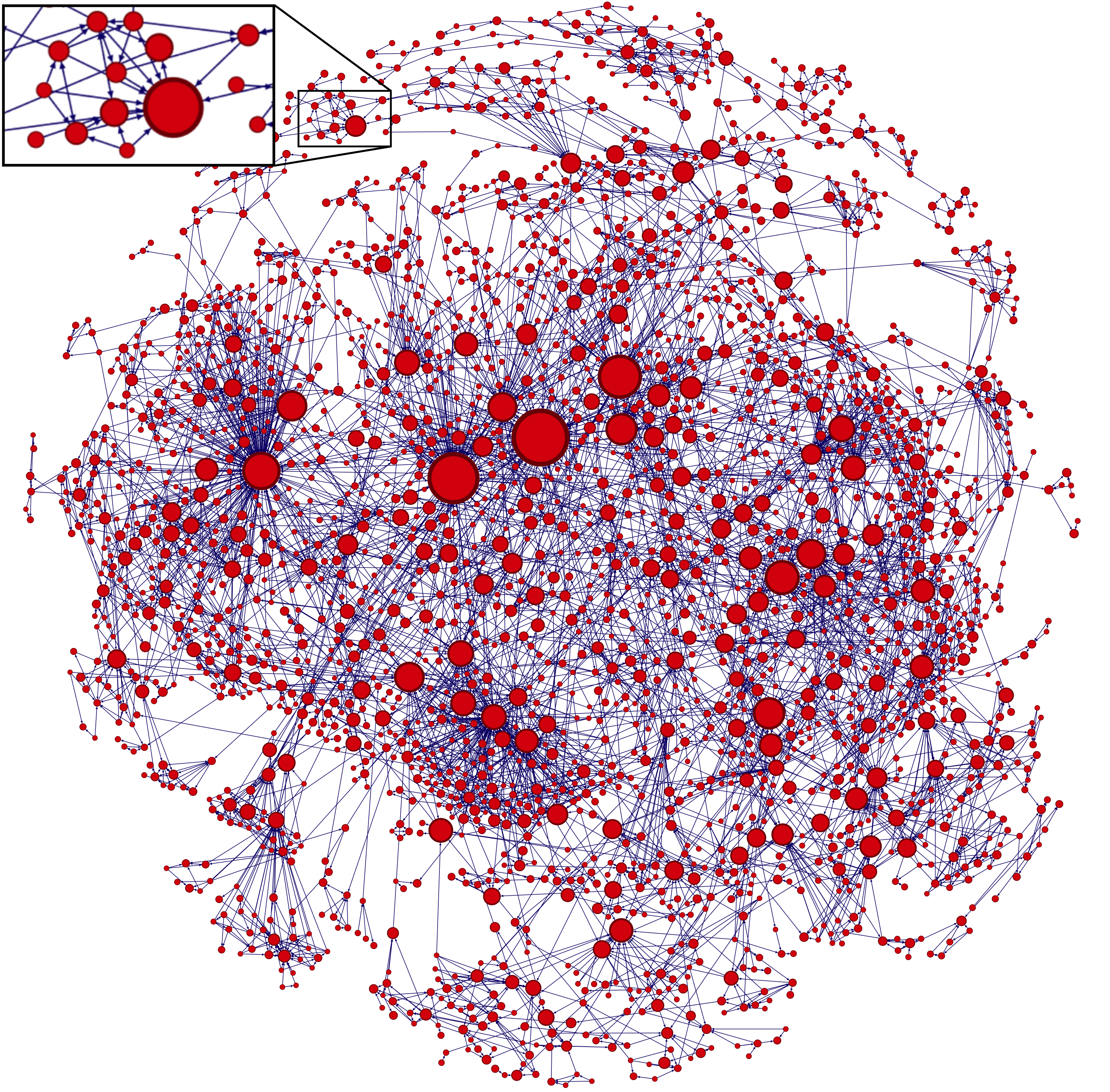}
 \caption{Visualization of the \textbf{Cora} graph based on embeddings learned from our gravity-inspired graph VAE model. In this graph, nodes are scaled using mass parameters $\tilde{m}_i$. The node separation is based on distances in the embedding space, using Force Atlas $2$ layout and the Fruchterman-Reingold algorithm \cite{fruchterman1991graph} on Gephi. Edges directionalities are best viewed on screen.}   
 \end{figure*} 
 
Table 2 reports the mean AUC and AP scores, along with standard errors over 100 runs, for each dataset and the three tasks. Training graphs and test sets are different for each of the 100 runs. Overall, our gravity-inspired models achieve competitive results.

On \textit{task 1}, standard graph AE and VAE models, despite ignoring the directionality, still perform reasonably well (e.g., $82.79\%$ AUC for standard graph VAE on Cora). This emphasizes the limited impact of the directionality on performances for such a task, as planned in Section 4.1.1. Nonetheless, our gravity-inspired models significantly outperform the standard ones (e.g., $91.92\%$ AUC for gravity-inspired graph VAE on Cora), confirming the relevance of capturing both proximity and directionality. Moreover, our models are competitive w.r.t. baselines designed for directed graphs. Among them, APP is the best on our three datasets, together with the source/target graph AE model on Google.

On \textit{task 2}, i.e., biased negative samples link prediction, our gravity-inspired models
achieve the best performances (e.g., a top $76.19\%$ AUC on Citeseer, $11+$ points above the best baseline). Models ignoring the directionality for prediction, i.e., node2vec and standard graph AE and VAE, totally fail ($50.00\%$ AUC and AP on all graphs, corresponding to the random classifier level), which was expected since test sets include both directions of each node pair. Experiments on \textit{task 3}, i.e., on bidirectionality prediction, confirm the superiority of our approach when dealing with tasks where directionality learning is crucial. Indeed, on this last task, gravity-inspired models also outperform the alternative approaches (e.g., with a top $85.82\%$ AUC for gravity-inspired graph AE on Google).

While the graph AE and VAE frameworks are based on different foundations, we found no significant performance gap in our experiments between (standard, asymmetric, or gravity-inspired) autoencoders and their variational counterparts. This result is consistent with previous insights from \cite{kipf2016-2, salha2019} on undirected graphs. Future work will investigate alternative prior distributions for graph VAE, aiming to challenge the traditional Gaussian hypothesis that, despite being convenient for computations, might not be an optimal choice in practice \cite{kipf2016-2}. Lastly, we note that all AE/VAE models required a comparable training time of roughly 7 seconds (respectively 8 seconds, 5 minutes) for Cora (resp. for Citeseer, for Google) on our machine. Baselines were faster: for instance, on the largest Google graph, 1 minute (resp. 1.30 minutes, 2 minutes) was required to train HOPE (resp. APP, node2vec).
\begin{figure*}[ht]
  \centering
  \subfigure[Task 1]{
  \scalebox{0.4}{\includegraphics{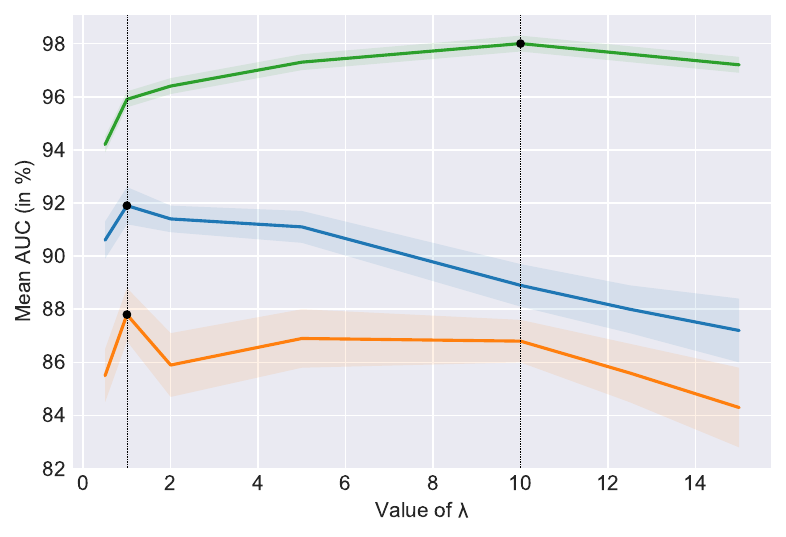}}}
  \subfigure[Task 2]{
  \scalebox{0.4}{\includegraphics{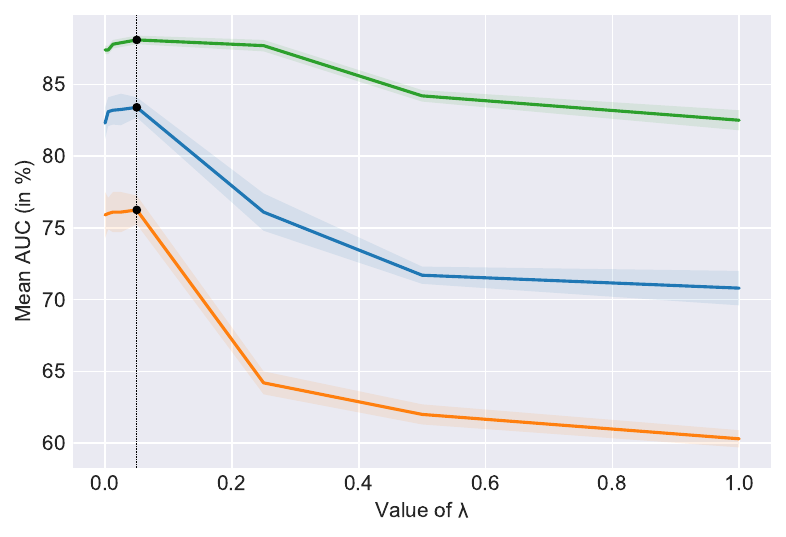}}}
  \subfigure[Task 3]{
  \scalebox{0.4}{\includegraphics{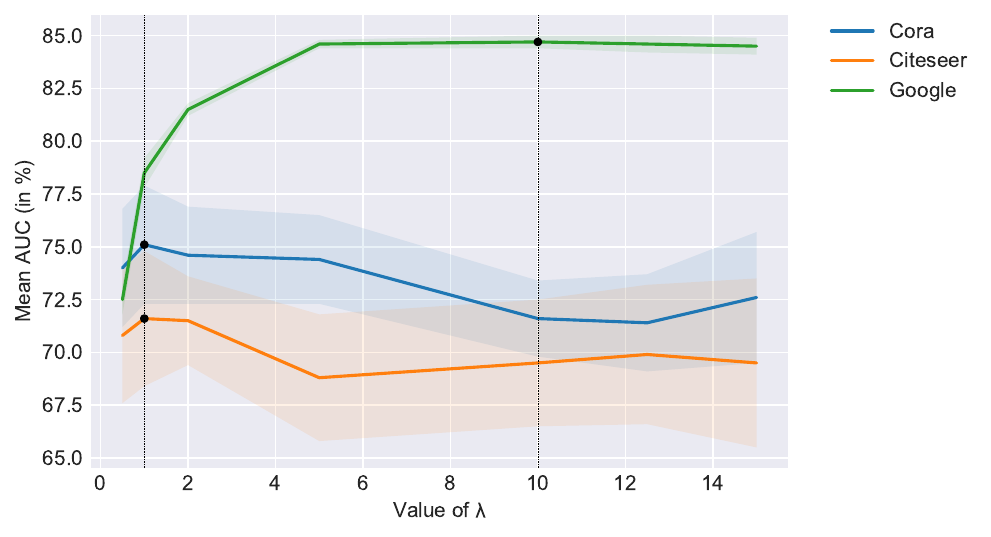}}}
  \caption{Impact of the parameter $\lambda$ on mean AUC, $\pm 1$ standard error, for gravity-inspired graph VAE.}
\end{figure*}

\subsection{Discussion}

To complete our experimental analysis, we propose a discussion on the nature of $\tilde{m}_i$, on the role of $\lambda$ to balance node proximity and influence, and on some extensions of our approach.

\subsubsection{Deeper insights on $\tilde{m}_i$}

Figure 1 provides a visualization of the Cora graph, using latent embedding vectors and $\tilde{m}_i$ parameters learned by our gravity-inspired graph VAE. In such a visualization, we observe that nodes with smaller "masses" tend to be connected to nodes with larger "masses" from their embedding neighborhood, which was expected due to our decoding scheme. 

From Figure 1, one might argue that $\tilde{m}_i$ tends to reflect centrality. To verify such a hypothesis, we compared $\tilde{m}_i$ to the most common graph centrality measures. Specifically, in Table 3, we report Pearson correlation coefficients of $\tilde{m}_i$ w.r.t. the following measures:

\begin{itemize}
    \item The \textit{in-degree} and \textit{out-degree} of the node, that are respectively the number of edges coming into and going out of the node.
    \item The \textit{betweenness centrality}, which is, for a node $i$, the sum of the fraction of all pairs shortest paths going through $i$: $c_B(i) = \sum_{s,t \in \mathcal{V}}\frac{sp(s,t|i)}{sp(s,t)}$, where $sp(s,t)$ is the number of shortest paths from node $s$ to node $t$, and $sp(s,t|i)$ is the number of those paths going through $i$ \cite{brandes2008variants}.
    \item The \textit{PageRank} score \cite{page1999pagerank},  ranking the importance of nodes based on the structure of the incoming links. PageRank was originally designed to rank web pages, and can be seen as the stationary distribution of a random walk on the graph~\cite{page1999pagerank}. 
    \item The \textit{Katz centrality}, a generalization of the eigenvector centrality. The Katz centrality of node $i$ is $c_i = \alpha \sum_{1\le j\le n} A_{ij} c_j + \beta$, where $A$ is the adjacency matrix with largest eigenvalue $\lambda_{\text{max}}$, usually with $\beta = 1$ and with $\alpha < \frac{1}{\lambda_{\text{max}}}$ \cite{katz1953new}.
\end{itemize}

\begin{table}[t]
\begin{center}
\begin{small}
\centering
\caption{Pearson correlation coefficient of centrality measures with parameter $\tilde{m}_i$, learned from our gravity-inspired graph VAE - Katz on Google not reported due to complexity.\label{tab:correlation_metrics}}
\begin{tabular}{c|c|c|c}
    \toprule
\textbf{Centrality Measures} & \textbf{Cora} & \textbf{Citeseer} & \textbf{Google} \\
\midrule
\midrule
\textbf{In-degree} & $0.5960$ &  $0.6557$ & $0.1571$\\
\textbf{Out-degree} & $-0.2662$ & $-0.1994$ & $0.0559$ \\
\textbf{Betweenness} & $0.5370$ &  $0.4945$ & $0.2223$\\
\textbf{Pagerank} & $0.4143$ & $0.3715$ & $0.1831$\\
\textbf{Katz} & $0.5886$ & $ 0.6428$ & - \\
    \bottomrule
\end{tabular}
\end{small}
\end{center}
\vspace{-0.5cm}
\end{table}

As observed in Table 3, the parameter $\tilde{m}_i$ is positively correlated with all of these centrality measures, except for the out-degree, where the correlation is negative (or almost null for Google), meaning that nodes with few edges going out of them tend to have larger values of $\tilde{m}_i$. Correlations are not perfect, emphasizing that our models do not learn one of these measures precisely. We also note that centralities are lower for Google, which might be due to the structure of this graph and especially to its density.

In our experiments, we tried to replace $\tilde{m}_i$ by any of these (normalized) centrality measures when performing link prediction, and to learn optimal vectors $z_i$ for these fixed masses values, achieving underperforming results. For instance, we reached an $89.05\%$ AUC by using betweenness centrality on Cora instead of the actual $\tilde{m}_i$ learned by the VAE, which is above standard graph VAE ($82.79\%$ AUC) but below the gravity-inspired VAE with $\tilde{m}_i$ ($91.92\%$ AUC). Also, using centrality measures as initial values for $\tilde{m}_i$ before model training did not significantly improve performances in our tests.

\subsubsection{Impact of the parameter $\lambda$}

In Section 3.5, we introduced a parameter $\lambda \in \mathbb{R}^+$ to tune the relative importance of the node proximity w.r.t. the mass attraction, leading to the reconstruction scheme $\hat{A}_{ij} = \sigma(\tilde{m}_j - \lambda \log \|z_i - z_j\|^2_2)$. In Figure 2, we show the impact of $\lambda$ on mean AUC scores for our VAE model and all three datasets. For Cora and Citeseer, on \textit{task 1} and \textit{task 3}, $\lambda = 1$ is the optimal choice, consistently with Newton's formula (see Figure 2 (a) and (c)). However, for Google, on \textit{task 1} and \textit{task 3}, we obtained better performances for higher values of $\lambda$, notably for $\lambda = 10$ that we used in our experiments. Increasing $\lambda$ reinforces the relative importance of the symmetric node proximity in the decoder, measured by $\log \|z_i - z_j\|^2_2$, w.r.t. the parameter $\tilde{m}_j$ capturing the global influence of a node on its neighbors and therefore asymmetries in links. Since the Google graph is denser than Cora and Citeseer, and has a higher proportion of symmetric relations (see Table 1), putting the emphasis on node proximity appears as a relevant~strategy.

On a contrary, on \textit{task 2} we achieved optimal performances for $\lambda = 0.05$, for all three graphs (see Figure 2 (b)). Since $\lambda < 1$, we, therefore, improved scores by assigning more relative importance to the mass parameter. This result is not surprising since, for the \textit{biased negative samples link prediction} task, learning directionality is more crucial than learning proximity, as node pairs from test sets are all included in both directions. As illustrated in Figure 2 (b), increasing $\lambda$ significantly deteriorates performances.

\subsubsection{Extensions and openings} 

Throughout these experiences, we focused on featureless graphs to fairly compete with HOPE, APP, and node2vec. However, as explained in Section 3, our models could easily leverage node features, in addition to the graph structure summarized in $A$. Moreover, the gravity-inspired method is not limited to GCN encoders and could be generalized to alternative graph neural networks. Our future research will provide more evidence on such extensions, will investigate better-suited priors for graph VAE, and will generalize the existing scalable graph AE/VAE framework \cite{salha2019} to directed graphs. We also aim to explore to which extent graph AE/VAE can tackle the node clustering problem in directed graphs.

\section{Conclusion}

In this paper, we presented a new method, inspired by Newtonian gravity, to learn node embedding representations from directed graphs using graph AE and VAE models. They effectively address challenging directed link prediction problems. Our work also pinpointed several research directions that, in the future, should lead towards the improvement of our approach.

\bibliographystyle{ACM-Reference-Format}
\bibliography{sample-base}

\end{document}